\documentclass[lettersize,journal, review]{IEEEtran}
\usepackage{amsmath,amsfonts}
\usepackage{array}
\usepackage[caption=false,font=normalsize,labelfont=sf,textfont=sf]{subfig}
\usepackage{textcomp}
\usepackage{stfloats}
\usepackage{url}
\usepackage{multirow}
\usepackage{verbatim}
\usepackage{graphicx}
\graphicspath{{Figures/}}
\usepackage[numbers,sort&compress]{natbib}
\usepackage{xcolor}
\usepackage[ruled,linesnumbered]{algorithm2e}
\usepackage{algpseudocode}
\usepackage{amssymb} 

\IEEEpubid{\begin{minipage}{\textwidth}\ \centering
		Copyright \copyright 2025 IEEE. Personal use of this material is permitted. \\
		However, permission to use this material for any other purposes must be obtained 
		from the IEEE by sending an email to pubs-permissions@ieee.org.
\end{minipage}}

\begin{document}

\title{Adaptive Depth-converted-Scale Convolution for Self-supervised Monocular Depth Estimation}

\author{Yanbo Gao, Huibin Bai, Huasong Zhou, Xingyu Gao, Shuai Li,~\IEEEmembership{Senior Member,~IEEE}, Xun Cai, \\Hui Yuan,~\IEEEmembership{Senior Member,~IEEE}, Wei Hua, Tian Xie
\thanks{Yanbo Gao and Xun Cai are with School of Software, Shandong University, Jinan 250100, China, and also with Shandong University-WeiHai Research Institute of Industrial Technology, Weihai 264209, China. \par
Huibin Bai, Huasong Zhou, Shuai Li and Hui Yuan are with School of Control Science and Engineering, Shandong University, and Key Laboratory of Machine Intelligence and System Control, Ministry of Education, Jinan 250100, China. E-mail:  shuaili@sdu.edu.cn.
\par Xingyu Gao is with the Institute of Microelectronics, Chinese Academy of
Sciences, Beijing 100029, China. 
\par Wei Hua and Tian Xie are with Research Institute of Interdisciplinary Innovation, Zhejiang Lab, Hangzhou, China.}
}



\maketitle
\begin{abstract}
Self-supervised monocular depth estimation (MDE) has received increasing interests in the last few years. The objects in the scene, including the object size and relationship among different objects, are the main clues to extract the scene structure. However, previous works lack the explicit handling of the changing sizes of the object due to the change of its depth. Especially in a monocular video, the size of the same object is continuously changed, resulting in size and depth ambiguity. To address this problem, we propose a Depth-converted-Scale Convolution (DcSConv) enhanced monocular depth estimation framework, by incorporating the prior relationship between the object depth and object scale to extract features from appropriate scales of the convolution receptive field.  The proposed DcSConv focuses on the adaptive scale of the convolution filter instead of the local deformation of its shape. It establishes that the scale of the convolution filter matters no less (or even more in the evaluated task) than its local deformation. Moreover, a Depth-converted-Scale aware Fusion (DcS-F) is developed to adaptively fuse the DcSConv features and the conventional convolution features. Our DcSConv enhanced monocular depth estimation framework can be applied on top of existing CNN based methods as a plug-and-play module to enhance the conventional convolution block. Extensive experiments with different baselines have been conducted on the KITTI benchmark and our method achieves the best results with an improvement up to 11.6\% in terms of SqRel reduction. Ablation study also validates the effectiveness of each proposed module.
\end{abstract}

\begin{IEEEkeywords}
Depth Estimation, Depth-converted-Scale Convolution, Depth-converted-Scale aware Fusion
\end{IEEEkeywords}

\section{Introduction}
\IEEEPARstart {D}{epth} {{estimation}} is a fundamental and important task in 3D vision, where the obtained depth maps are used in various {{downstream tasks}} including augmented reality, autonomous driving and 3D reconstruction by providing useful scene information \cite{10325550,8764412,li2023depthformer,10261254,lei18depth,8693882}. While the  {{depth acquisition}} hardware such as LiDAR has improved in the last few years, they still suffer from some problems including the sparse depth points and high costs.{{ Therefore, depth estimation methods using color image/video have been widely studied. Especially with the rapid development of deep learning, image/video based depth estimation using deep learning has been actively investigated \cite{eigen2014depth,bhat2021adabins, godard2019digging,LiftFormer2025lift}, including both supervised and self-supervised learning methods.}} 

\begin{figure}[t]
\centering
\includegraphics[width=0.5\textwidth]{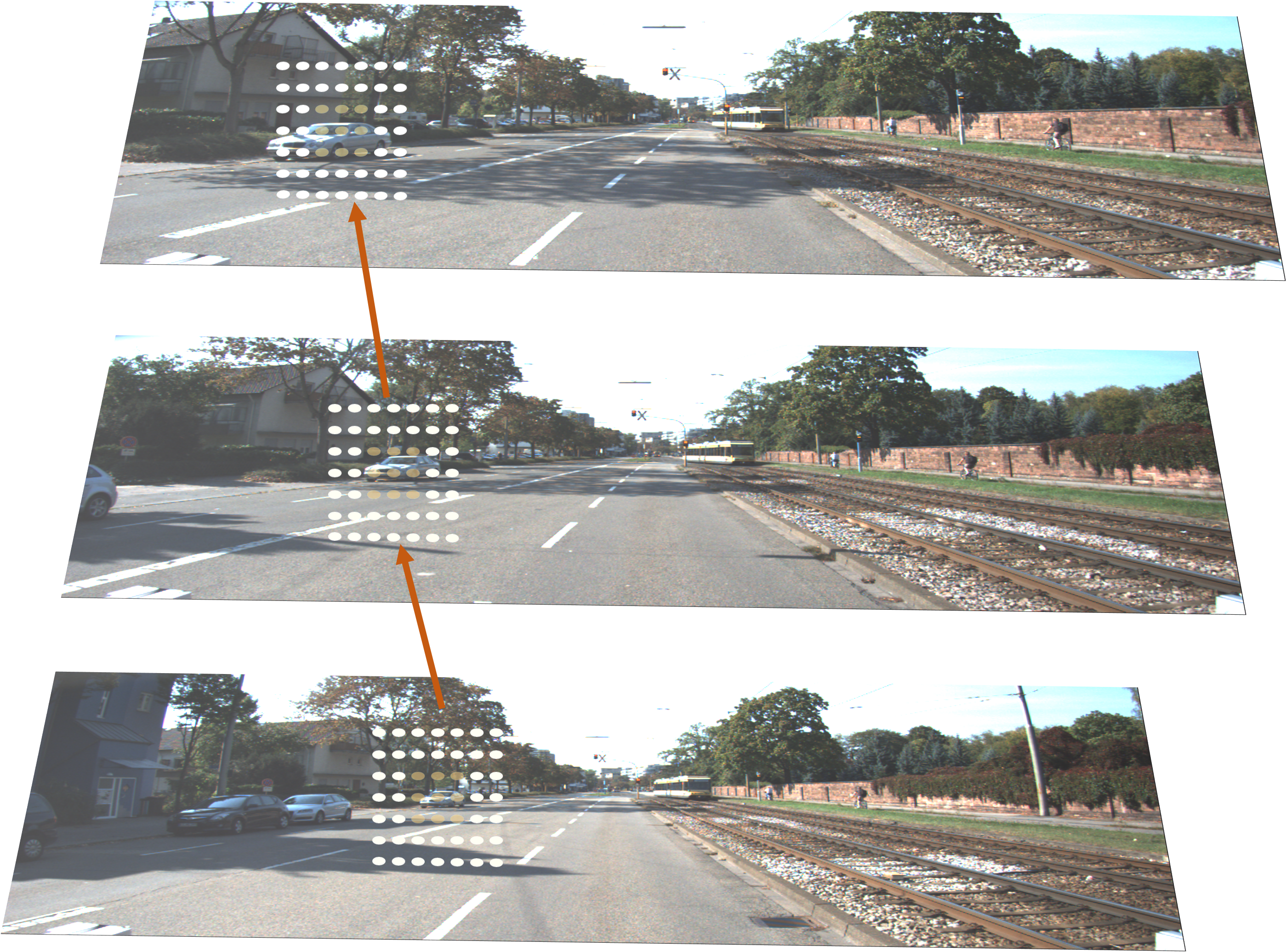}
\caption{Illustration of the object size and depth change at successcive frames of a monocular video. }
\label{fig_motivation}
\end{figure}

Supervised depth estimation methods \cite{eigen2014depth,bhat2021adabins} use ground truth depth maps for supervision, where a large dataset with accurate depth labels is needed. Such a large dataset is usually difficult and expensive to obtain \cite{godard2019digging}. As an effective alternative, self-supervised methods have been proposed using a monocular video \cite{zhou2017unsupervised,zhou2021r,10348603,10711868} or stereo image pairs \cite{godard2017unsupervised,tosi2019learning,peng2021excavating,li2022self}. One image is used as input and the other image (the other view or a neighboring frame) is used as the target. The reconstruction error of synthesizing the target image with the estimated depth map is used as the training objective. In such a case, ground truth depth map is not required, making the training process much simpler. This paper focuses on the monocular video-based depth estimation.

The most important thing in depth estimation is to obtain the overall structural information of a scene including the relative depths of the different objects in the scene \cite{yan2021channel}. \IEEEpubidadjcol  To obtain the global structural information and provide dense depth map estimation, an encoder-decoder architecture is usually adopted \cite{godard2019digging,li2022self,yan2021channel}, where pooling operation (down-sampling) is used in the encoder to obtain the global scene information and deconvolution (up-sampling) is used in the decoder to produce dense depth maps. Image/feature pyramid is also used in \cite{song2021monocular} to facilitate the extraction of multi-scale features in order to accommodate the different sized objects. {However, these fixed scale changes in the pooling and pyramid cannot provide flexible local receptive fields to fully explore the varying and flexible sizes of the objects in the scene.} Especially for the depth estimation of a video captured with a moving camera, the sizes of the objects are continuously changed. In such cases, the existing convolutional networks with fixed-size convolution and fixed-scale processing cannot effectively explore the objects of different sizes and extract the 3D scene structural information. {As shown in Fig. \ref{fig_motivation}, the object size and depth are continuously changed among successive frames and the same receptive field covers different parts of the car.} Moreover, the relationship between the object depth and object scale is ignored, which inevitably degrades the extraction of object features at different depths of a scene.

In addition, most of the efforts on the general CNNs based computer vision methods have been devoted to the local deformation of the filter shape such as the deformable convolution \cite{dai2017deformable1}, instead of the scale of the convolution. More importantly, it seems to be a default setting that the scale of convolution over all locations of the feature map is fixed in each process and multiscale processing can only be obtained by predefining different convolution layers. The importance of adaptive scale at each location of the convolution is completely ignored.

To address the above problems, we propose a Depth-converted-Scale Convolution (DcSConv) by incorporating the prior relationship between the object size and its scene depth into determining the scale of the convolution. For a monocular video, the size of an object in the image changes continuously among the successive frames and is inversely proportional to its depth. To deal with this continuing size changes of the same object in different frames and receive information from a receptive field of appropriate scale, an adaptive scale convolution is developed based on the depth with the above depth-to-scale conversion relationship. Specifically, two ways of adaptive scale convolutions are developed to obtain scale-aware features, including Depth–converted Multiple Scale convolution Fusion (DMSF) to adaptively fuse multi-scale features based on the scale information, and Depth-converted-Scale Convolution (DcSConv) to directly perform convolution with a depth converted scale. The DcSConv can be generalized to a learned-scale convolution, focusing on the flexible scale of the convolution instead of the local deformation the filter. {In this way, the features can be extracted adaptively at different scales for objects at different depths, better capturing the details of objects and also the relationship of the same object among different frames.} Then, a depth-converted-scale aware fusion (DcS-F) module is developed to adaptively fuse the scale-aware features and the conventional convolution features according to the scale information. {By combing these two features, the relationship among objects within the same frame and the relationship of the same object among different frames can be both captured, thus improving the depth estimation performance.} 

The main contributions of this paper can be summarized as follows.

\begin{itemize}
\item{We propose an adaptive scale convolution based on the depth information and the depth-to-scale conversion relationship. It solves the problem of size ambiguity of the same object at different depths among different frames and forms receptive fields of appropriate scales. To the best of our knowledge, this is the first work that investigates the adaptive scale of the convolution, instead of local deformation of the convolution filter. Experiments also validate the effectiveness of our adaptive scale convolution over the rather locally deformable convolution.}
\item{We develop two adaptive scale convolution methods, namely, the Depth–converted Multiple Scale convolution Fusion (DMSF) and the Depth-converted-Scale Convolution (DcSConv). The latter DcSConv can be generalized to learned-scale convolution (GLSConv), to directly learn appropriate scale from the features.}
\item{We design a Depth-converted-Scale aware Fusion (DcS-F) module to adaptively fuse the scale-aware features and the conventional convolution features based on the scale information.}
\end{itemize}

The proposed DcSConv and DcS-F do not concern the detailed architecture of the monocular depth estimation network and can be used to replace the basic convolution in any existing networks. Experiments on two baseline networks \cite{godard2019digging,yan2021channel} are conducted and the proposed method achieves the best performance. Ablation studies are also performed to validate the effectiveness of each proposed module.

\section{Related Work}
\label{sec_related}
In this {section}, we present a brief review of the related studies for deep learning based monocular depth estimation, including the supervised depth estimation methods \cite{laina2016deeper, fu2018deep, ranftl2021vision, patil2022p3depth, kendall2018multi, wu2022fast, ranftl2020towards} and self-supervised ones \cite{mahjourian2018unsupervised, wang2021can, yin2018geonet, hui2022rm, petrovai2022exploiting, choi2021adaptive, jung2021fine, ji2021monoindoor}. Moreover, some related works with multi-scale feature processing \cite{zhao2017pyramid, chen2020dynamic, ding2020learning} are also briefly described.

\subsection{Supervised Depth Estimation}
The goal of the supervised depth estimation is to predict the scene depth map from RGB images with the ground truth for supervision. Generally, the CNN based encoder-decoder architecture is used to extract features from the image via the encoder and provide the depth map as output via the decoder. Based on this backbone architecture, many works on improving encoder or decoder architectures have been developed. Eigen et al. \cite{eigen2014depth} proposed a two-stage framework which further refines the coarse depth map generated from the encoder-decoder network based on the local information. The residual network is adopted in \cite{laina2016deeper} as an encoder to better extract scene information. The dilated convolution is used instead of the standard convolution in \cite{fu2018deep} to increase the receptive field. The transformer-based block is also investigated in \cite{ranftl2021vision} to improve the feature extraction in the encoder.

In addition to using different deep learning layers, different architectures have also been explored. In \cite{song2021monocular}, Laplacian pyramid is used to process the input with Laplacian residuals of different scales as inputs to compensate the upsampling errors in the encoder-decoder architecture on the corresponding predicted depth maps. To enhance the global scene structure of predicted depth, the piecewise planarity prior is embedded into the network design which enables network to selectively leverage information from coplanar pixels \cite{patil2022p3depth}. Cipolla et al. \cite{kendall2018multi} proposed to perform semantic segmentation and depth estimation simultaneously and use scene semantic information to obtain shaper object boundaries for depth estimation. Piccinelli et al. \cite{piccinelli2023idisc} proposed an Internal Discretization module which implicitly partitions the scene into a set of high-level patterns for depth estimation. Yuan et al. \cite{yuan2022new} proposed Neural Window Fully-connected Conditional Random Fields to capture the relationship using a graph to further improve the depth prediction performance. These supervised methods generally achieves better performance than self-supervised methods, with the cost of ground truth depth map.  
\vspace{0.3cm}

\subsection{Self-supervised Monocular Depth Estimation}
To overcome the constraints of using the ground truth depth map in the supervised depth estimation methods, self-supervised monocular depth estimation methods have been studied, which use the reconstruction error of synthesizing the neighboring frame in a monocular video with the estimated depth map as supervision. Zhou et al. \cite{zhou2017unsupervised} first jointly trained a depth estimation network and a pose prediction network to conduct the synthesis among frames.  Mahjourian et al. \cite{mahjourian2018unsupervised} further proposed to align the point clouds generated from adjacent estimated depth maps and designed a 3D point cloud alignment loss based on the ICP algorithm to provide 3D geometric constraints. In such synthesis based supervision, the pose and the depth are used together, which may result in scale difference among different frames when the pose is estimated differently. To reduce this problem, a scale aware geometric loss is introduced in \cite{wang2021can}, to enforce scale consistency by using a point cloud alignment constraint. Godard et al. \cite{godard2019digging} proposed the per-pixel minimum reprojection loss from two neighboring frames to deal with the occlusion pixels and an auto-masking to filter out the moving targets pixels. Yin et al. \cite{yin2018geonet} adopted the rigid structure reconstructor and non-rigid motion localizer to estimate static scene geometry and capture dynamic objects separately. The occlusions and non-Lambertian surfaces are further detected with the predicted optical flow in the non-rigid motion localizer. Hui et al. \cite{hui2022rm} also processed the camera and object motions separately and constrained them by image warping with both object motion and depth projection. Residual pose estimation module was designed in \cite{ji2021monoindoor} to alleviate inaccurate pose prediction and in turn improve the depth estimation. The pose network is firstly used to predict an initial camera relative pose and then the residual pose network is used to further iteratively predict residual camera poses. 

Other than using synthesis based reconstruction for supervision, there are also works exploring other ways to generate depth maps for supervision or using other related information for auxiliary supervision. Petrovai et al. \cite{petrovai2022exploiting} used a self-distillation based training pipeline to first generate high-resolution depth maps as pseudo labels and then re-train the network to fit both low- and high- resolution images to increase scale consistency. Depth maps generated from the conventional semi-global matching algorithm or self-supervised stereo matching method of stereo images pairs are used as the additional auxiliary supervision in \cite{choi2021adaptive}.  In \cite{jung2021fine}, the semantic boundary information is used to constrain depth decoder features via a triplet loss to enhance the depth representation around the object boundary.

\subsection{Multi-Scale Feature Extraction in Depth Estimation }

It is known that multiple scale features such as feature pyramid are important to characterize an image or for down-stream tasks since one object can be of different scales in different images. There also exist multi-scale feature extraction methods in deep learning based depth estimation to accommodate the different sized objects. Typical multiscale processing techniques includes the general pooling and up-sampling (deconvolution) in the encoder-decoder architecture, image and feature multiscale processing such as image pyramid \cite{song2021monocular} and spatial pyramid pooling \cite{zhao2017pyramid}, and different convolutions such as dilated convolution \cite{fu2018deep}. While they can achieve multi-scale processing to some extent, only a few (usually three or four) predefined fixed-scale changes are obtained with predefined fixed-size convolution. In the general computer vision tasks other than depth estimation, there also exist several ways to change the convolution receptive field to extract features of different scales. Dynamic convolution \cite{chen2020dynamic} was proposed to aggregate features generated from multiple convolution kernels, where an attention mechanism is used to generate weights to combine the features from different convolutions. In \cite{ding2020learning}, depth guided convolution was proposed using different dilated convolutions and combining them based on an adaptive weight obtained by the image feature. Depth is used to generate dynamic convolution filter weights instead of scales.  Deformable convolution \cite{dai2017deformable1} can also, to some extent, adapt the receptive filed by changing the sampling gird with learned offsets. However, it focuses more on the local deformation instead of processing with different scales, and its offsets generally fall in a small neighborhood around the optical flow as illustrated in the deformable offset diversity \cite{ding2020learning}. Compared to these methods, our work develops a Depth-converted-Scale Convolution to achieve adaptive-scale processing based on the relationship between the object size and its depth, which theoretically stands. Especially in the monocular video-based depth estimation where the same object is of continuous scale change among different frames due to the change of its depth, our contribution becomes even critical.

\begin{figure*}[!t]
\centering
\includegraphics[width=0.9\textwidth]{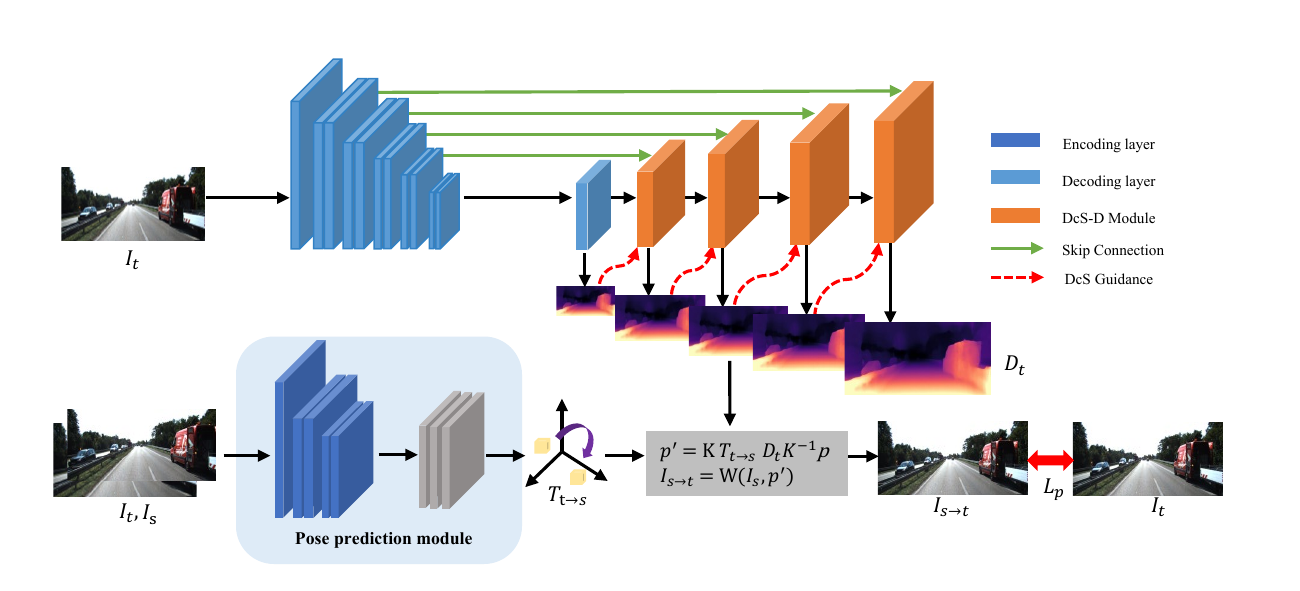}
\caption{The framework of the proposed adaptive Depth-converted-Scale Convolution (DcSConv) enhanced monocular depth estimation. The progressively generated depth maps of different resolutions are used to further update the network by constructing adaptive-scale convolution based on the depth value. }
\label{fig_1}
\end{figure*}

\section{Proposed Method}
\label{sec_proposed}
\subsection{Overview}
\label{sec_ow}
Given a monocular video, denote the target frame by $I_{t}$  and its corresponding to-be-predicted depth map by $D_{t}$, and the neighboring frames by the source frames $I_{s}$ (generally using two adjacent frames with $ s = \left \{ t-1, t+1 \right \}$). The goal of self-supervised monocular depth estimation is to predict the depth map $D_{t}$ from the target frame $I_{t}$, supervised by the reconstruction error of synthesizing the neighboring frame with the predicted depth map (or synthesizing the current frame with the predicted depth map from backward warping). For a monocular video, the camera relative pose $T_{t{\rightarrow}s}$  between the target frame $I_{t}$ and the source frame $I_{s}$ also needs to be estimated for the synthesis process. Therefore, the whole network framework consists of a depth estimation module and a pose prediction module as shown in Fig. \ref{fig_1}.

For a continuous video captured with a moving camera, the sizes of the objects in a video are continuously changed. To accommodate this continuous varying object size changes, an adaptively scaled convolution is developed based on the prior knowledge between the object size and its relative depth in the scene. Generally speaking, for an object farther in the scene with a larger depth, the object size becomes smaller in the image and the corresponding size of convolution filter is also supposed to be smaller to match among different scales. On the other hand, depth estimation networks \cite{godard2019digging,li2022self,yan2021channel} generally adopt a sequential decoding architecture, where depth maps of different resolutions are progressively obtained, which makes the low-resolution depth maps available in the estimation process. Moreover, an initial depth map can always be obtained by a pre-trained model. Therefore, an initial or up-sampled depth map can be used to progressively guide the depth estimation process.  Conventionally, the depth map is not further explored by the network or only used as input (in the form of hidden features) to the next layer. In this paper, we further explore the depth map to update the local architecture of the following networks by projecting the different depths to different scaled convolution. 

The overall architecture of our self-supervised monocular depth estimation network is shown in Fig. \ref{fig_1}. An encoder-decoder architecture is used as the backbone as in \cite{godard2019digging,yan2021channel,ling2021unsupervised}. A Depth-converted-Scale aware feature Decoding (DcS-D) module with Depth-converted-Scale Convolution (DcSConv) is developed to adapt geometric scale changes of objects and fully extract 3D scene structural information. Our method does not concern the detailed network architecture, rather incorporating depth-based scale information into the network. Thus, any existing encoder-decoder architecture-based networks can be used with any type of encoder. In our experiments, Monodepth2 \cite{godard2019digging}, CADepth \cite{yan2021channel} and MonoviT \cite{zhao2022monovit} are used and verified as the baselines.

\begin{figure}[!t]
\centering
\includegraphics[width=0.3\textwidth]{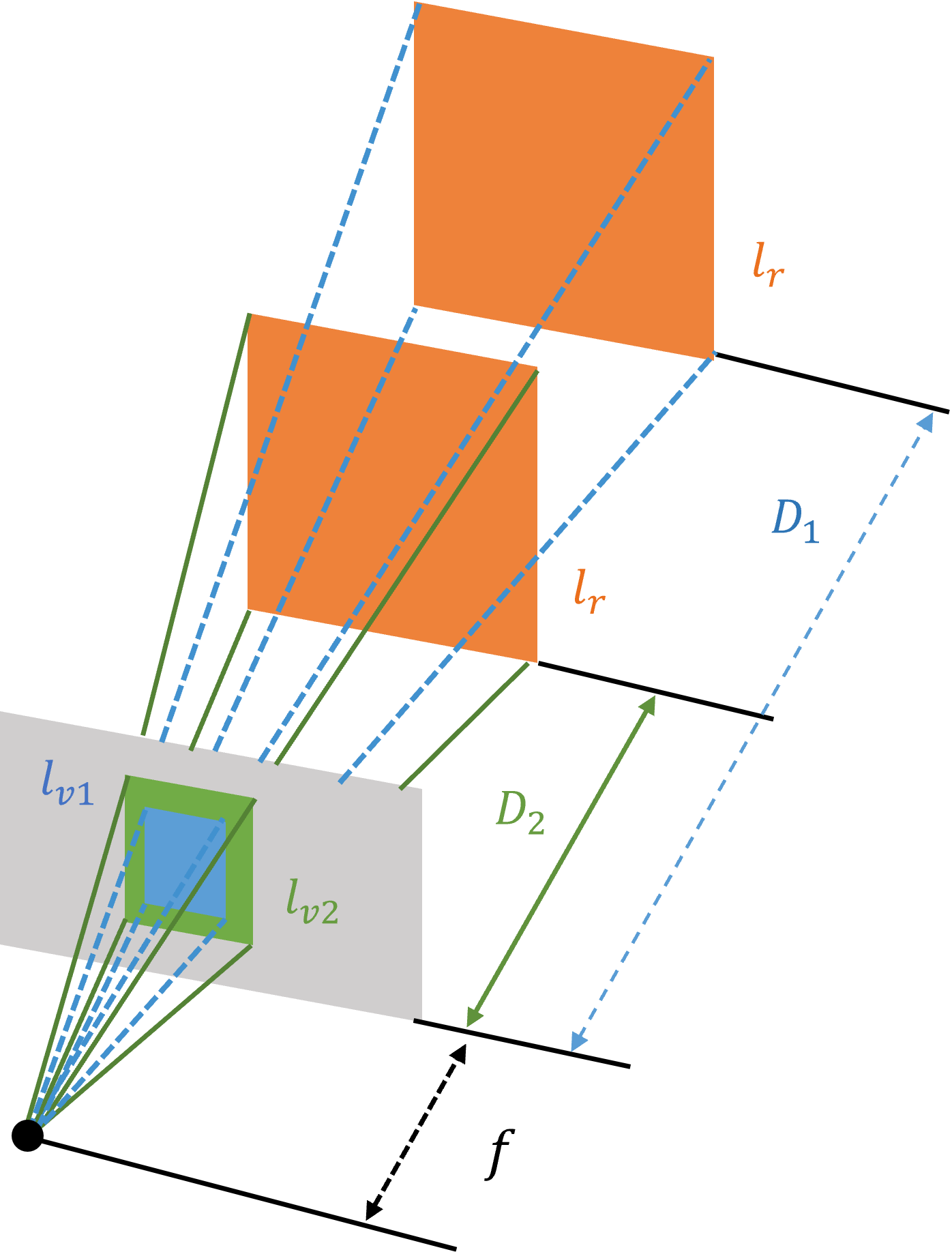}
\caption{Illustration of the relationship between the object scale in an image and its scene depth. { The length $l_r$} in the 3D is projected to different scales in an image with different depths. }
\label{fig_2}
\end{figure}

\subsection{Depth-scale Conversion}
\label{sec_dc}
The relationship between object size in an image and its scene depth is first investigated to provide the basis for DcSConv. { Specifically, considering the convolution performs the processing at each pixel location with the sliding window operation to capture the characteristics at each location, we focus more on the center pixel at each processing.  Therefore, a square centred at each pixel in the image, corresponding to a local receptive field, and a square at the object point projected to the center pixel, corresponding to an area centred around the object point in the 3D scene, are investigated. For simplicity, the length (height/width) of the square in the 3D scene and length in the image are investigated.} As shown in Fig. \ref{fig_2}, the relationship between the {lengths in the 3D scene and in the image} can be obtained under the ideal imaging principle of the pinhole camera. The {length} in the image can be obtained by
\begin{equation}
\label{equ_1}
{{l}}_{v}={{l}}_{r} \frac{f}{D} 
\end{equation}
where {$l_{r}$} and {$l_{v}$} denote the {lengths} in the 3D scene and in the image, respectively.  $D$ and $f$ denote its scene depth and camera focal length, respectively.

\begin{figure*}[!t]
\centering
\includegraphics[width=0.95\textwidth]{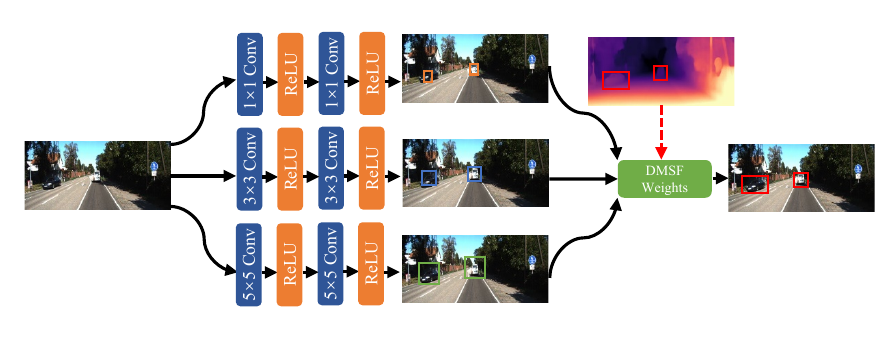}
\caption{Illustration of the Depth-converted Multiple Scale convolution Fusion (DMSF). }
\label{fig_3}
\end{figure*}

When the depth of the object is changed from $D_{1}$ to $D_{2}$, the {length} in the image is changed as follows.
\begin{equation}
\label{equ_2}
{{l}}_{v2}={{l}}_{v1} \frac{D_{1}}{D_{2}} 
\end{equation}
where $l_{v1}$ and $l_{v2}$ denote the {lengths} in the image under two different depths  $D_{1}$ and $D_{2}$, respectively. It can be observed that the {length} in the image is inversely proportional to its depth.

For the monocular video which is collected by the moving camera, the object depth is continuously changed among different frames. Thus, the same object in different images is of different sizes. To effectively capture the features of the object and make them consistent among different frames, it is better to capture the features of the object with a same scale in the image. That is to say, with changed depth and changed scale of the object in the image, the scale of the convolution filter is better to be changed accordingly. Assume the basic convolution filter size is $k_{r} \times {k_{r}}$  corresponding to { a reference} depth $D_{r}$, for the object at any depth $D$, the length of its convolution filter size can be obtained as
\begin{equation}
\label{equ_3}
{{k}}_{d}={{k}}_{r} \frac{D_{r}}{D} 
\end{equation}

Since existing networks commonly use a convolution filter size of  $3 \times 3$, the basic convolution filter size $k_{r} \times {k_{r}}$ is set to $3$ in this paper. The reference depth $D_{r}$ can be obtained as the mean depth of the scene or empirically set, which is validated in the experiments. Therefore, an appropriate convolution filter size can be obtained with the depth of the object, which can be determined with an initial depth map by a pre-trained model or up-sampled depth map by a previously decoded low-resolution depth map as discussed in subsection \ref{sec_ow}.

\subsection{Depth-converted Scale-aware Convolution Block}
\label{sec_db}
\subsubsection{Depth–converted Multiple Scale convolution Fusion (DMSF)} 

To obtain adaptive features of different scales at different locations according to their depths, the straightforward way is to extract multi-scale features first and then fuse them according to their depth-converted scales. As shown in Fig. \ref{fig_3}, three parallel branches with different convolution filter sizes are first used to obtain multi-scale features. Since the feature $F$ is usually generated using $3 \times 3$ standard convolution as in the existing methods, $3 \times 3$ convolution filter is used and supplemented by larger and smaller filters of $1 \times 1$ and  $5 \times 5$, respectively. Then the features extracted with different scales are fused to obtain depth-converted scale features. Instead of adaptively fusing them such as channel attention, the prior knowledge on the depth-converted scale is injected into the fusion process.  Specifically, the gaussian distance function is used to measure the distance between the depth-converted scale and the filter size used in the different branches, and then softmax is used as the normalization function to obtain the weight $a_{i}$ of each branch for fusion as 
\begin{equation}
\label{equ_4}
{{a}}_{i}=softmax(e ^ {- \frac {{\left|{k_{d}-k_{i}}\right|} ^ 2} {2 {\sigma} ^ 2} }),   i {\in} \left[ {1, 2, 3} \right]
\end{equation}
where $k_{d}$ is the depth-converted scale and $k_{i}$  represents the convolution filter size of each branch. $\sigma$ is a parameter used to normalize the scale difference and control the weight distribution among different branches. A larger $\sigma$ makes the process close to an average fusion over all branches, while a smaller $\sigma$ makes the process close to a nearest neighbor selection with a larger weight on the branch close to the depth-converted scale.  It is empirically set to $10$ in the experiments.

Finally, the output feature $F_{d}$ is generated by fusing features from three branches with the corresponding weights:
\begin{equation}
\label{equ_5}
{{F}}_{d}= \sum_{i=0} ^ 3 {a_{i} F_{i}}
\end{equation}

Since the fusion weights $a_{i}$ indicate the distance between the depth-converted scale and the filter size of each branch, this fusion process greatly enhances the capability of extracting dynamic-scale features based on the object size changes.

\begin{figure*}[!t]
\centering
\includegraphics[width=0.78\textwidth]{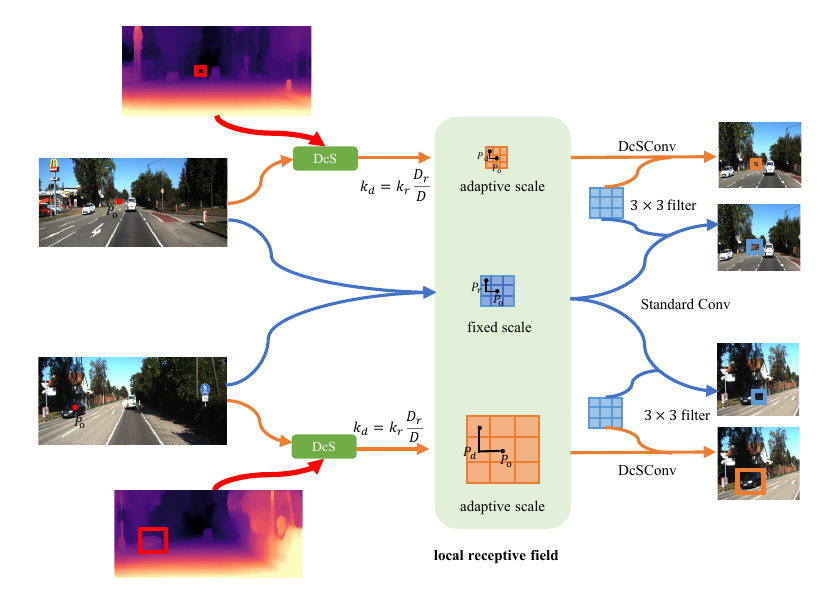}
\caption{Illustration of  Depth-converted-Scale Convolution (DcSConv) in comparison with the standard convolution among the successive frames. ``$P_o$'' denotes each location on the input feature map and the center of each convolution operation, and ``$P_r$'', ``$P_d$'' denote the sampling locations in the local grid, in fixed receptive field as the conventional convolution and the depth-converted-scale receptive field in the proposed DcSConv, respectively. With the object scale information provided by the corresponding depth map, our DcSConv can adaptively capture object scale changes with adaptive local receptive field.}
\label{fig_4}
\end{figure*}

\subsubsection{Depth–converted-Scale Convolution (DcSConv)} 

DMSF first extracts multi-scale features and then fuses them to achieve a depth-converted-scale convolution feature. In this case, the depth-converted-scale feature  is interpolated by multiple fixed-scale features, which inevitably brings scale distortion to the feature. In this section, a Depth-converted-Scale convolution (DcSConv) is developed, which directly produces a flexible-scale convolution feature based on the depth.

Generally, 2D standard convolution moves the convolution kernel sequentially on the input feature map with the sliding window operation until the convolution kernel covers the entire input. As shown in Fig. \ref{fig_4}, for each location $P_{o}$  on the input feature map $F$, the 2D convolution consists of the two steps: 1) performing sampling operations on the input feature map $F$ with a regular grid $R$ known as the receptive field; 2) summing the sampled feature values with the corresponding convolution filter weights $w$. For the corresponding location on the output feature map $F_{c}$, the convolution output can be obtained as:
\begin{equation}
\label{equ_6}
F_{c}(P_{o}) = \sum_{P_{r} \in R_{c}} w(P_{r}) \cdot F(P_{o} + P_{r})
\end{equation}
where $R_{c}$ indicates the local grid and $P_{r}$ enumerates the sampling locations in $R_{c}$. {$w(P_{r})$ are the learnable weights.} For a standard convolution using a convolution filter size $k_{r} = 3$ for example, $R_c$ can be defined as:
\begin{equation}
\label{equ_7}
R_{c} = {(-1,-1), (-1,0), ..., (0,1), (1,1)}
\end{equation}

To effectively capture the object size changes due to the depth change among monocular successive frames, as described in {the subsection} \ref{sec_dc}, a depth-converted-scale $k_{d}$ is obtained and the receptive filed of the DcSConv can be adaptively adjusted on different objects according to their depths. The DcSConv at each location  $P_{o}$ becomes:
\begin{equation}
\label{equ_8}
F_{DcSC}(P_{o}) = \sum_{P_{d} \in R_{DcSC}} w(P_{d}) \cdot F(P_{o} + P_{d})
\end{equation}
where {$w(P_{d})$ are the learnable weights of DcSConv and can be learned in the same way as the standard convolution.} $ R_{DcSC}$ represents the depth-converted-scale grid $k_{d} \times {k_{d}}$ and $P_{d}$ enumerates the locations in $R_{DcSC}$. For simplicity, as shown in Fig. \ref{fig_4}, same sampling points and sampling strategy as the conventional convolution are used in each convolution, i.e., $9$ points are sampled including the center, the corners and the center of edges as the conventional $3 \times 3$ convolution. Therefore, $R_{DcSC}$ can be expressed as
\begin{equation}
\label{equ_9}
R_{DcSC} = {(-\frac {k_{d}-1} {2},-\frac {k_{d}-1} {2}),..., (0,0)..., (\frac {k_{d}-1} {2},\frac {k_{d}-1} {2})}
\end{equation}

In our DcSConv, the sampling location $P_{d}$  in $R_{DcSC}$ may not lie on integer locations rather on fractional ones. Here, the bilinear interpolation among features is used to generate the features at the fractional positions. Fig. \ref{fig_4} illustrates the difference between our DcSConv and the standard convolution, where our DcSConv can receive information from adaptive receptive fields to adapt to different scales of the same object at different depths. 

\subsubsection{Generalized Learned-Scale Convolution (GLSConv)} 

The proposed DcSConv can be generalized as a learned-scale convolution, consisting of two parts: learning depth from features and transfer depth to scale. The final scale can be expressed as 
\begin{equation}
\label{equ_10}
k_{d} = f_{d-s} (f_{f-d} (F_{in}))
\end{equation}
where $F_{in}$ is the input feature to the GLSConv layer, $f_{f-d}$ and $f_{d-s}$ represent the learned feature to depth transformation and the depth to scale transformation, respectively. $f_{f-d}$ can be any learnable network such as the general depth estimation module, while $f_{d-s}$ is the depth converted to scale process described in the {subsection} \ref{sec_dc}.

More generally, the fixed depth to scale transformation $f_{d-s}$ can also be generalized to a learnable module such as a linear layer to learn the transformation from depth to a flexible convolution scale. This is also validated in the experiments where it improves the performance over the fixed-scale convolution. However, as described in {the subsection} \ref{sec_dc}, there is a direct mathematical transformation relationship existing between the depth and scale, thus a learnable one may not outperform the DcSConv (detailed in the Ablation Study). In addition, with $f_{f-d}$ and $f_{d-s}$ both being learnable transform, they can be merged to an overall learnable network as $k_{d} = f_{f-s}(F_{in})$, where $f_{f-s}$ represents the overall transformation from input feature to the convolution scale. This GLSConv can be theoretically applied in any CNNs based methods.

Currently, CNNs focus more on the deformation of the filter shape such as the deformable convolution, instead of the convolution scale. Note that while the convolution scale can also be represented as a deformation of convolution filters, the deformable convolution works mostly in the form of local filter shape transformation and unable to obtain large scale convolution. Most of the scale processing, such as pyramid, dilated convolution, pooling, are predefined and fixed over the whole image. This paper demonstrates that the scale of the convolution also matters significantly as validated in our experiments that the proposed DcSConv performs better than the deformable convolution in the depth estimation task.

\subsection{Depth-converted-Scale aware Fusion(DcS-F) Module}
\begin{figure}[!t]
\centering
\includegraphics[width=0.5\textwidth]{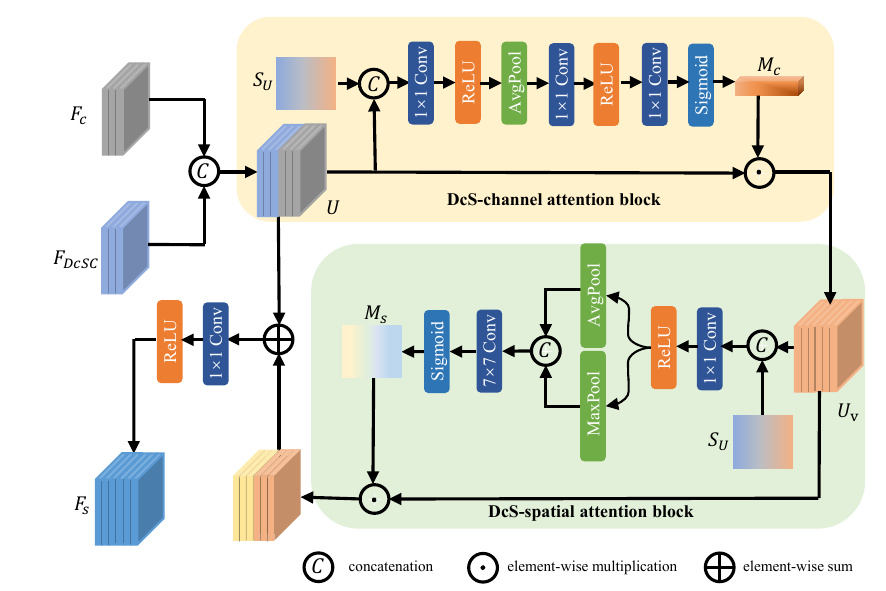}
\caption{Illustration of the Depth-converted-Scale aware Fusion (DcS-F) module. It consists of a DcS-channel attention block and a DcS-spatial attention block based on the depth information to fully fuse the DcSConv features and conventional convolution features. }
\label{fig_5}
\end{figure}

With our DcSConv, the capability of modeling object geometric transformations especially in terms of scale is greatly enhanced in the scene depth changing scenarios. While our DcSConv solves the problem of scale ambiguity under different depths among different frames, the features of different objects within a frame at the same convolution scale also contains important information of the scene. {It indicates the relationship between the different objects at a same frame. Moreover, the features obtained at the same scale can also serve as the anchor information to help process the DcSConv features.} Therefore, a Depth-converted-Scale aware Fusion (DcS-F) module is further developed to fuse the DcSConv features and the conventional convolution features according to the depth/scale information. Considering that different locations of the features are obtained with convolutions of different sizes based on their scales in the DcSConv features, to adaptively fuse with the conventional convolution features, the proposed DcS-F is developed with a DcS-channel attention block and a DcS-spatial attention block, as shown in Fig. \ref{fig_5}.

Specifically, the DcSConv feature $F_{DcSC}$  and the conventional convolution feature $F_{c}$ are firstly concatenated as the initial feature $U$. The DcS-channel attention block first fuses the features in a channel-wise manner to obtain the global channel weights. {The scale map is normalized to a scale difference map $S_U$,  by first subtracting the depth converted scale $k_d$ with the reference convolution filter size $k_r$ (3), then divided by $3$. This normalization process $(k_d-3)/3$ aims to make the scale difference map roughly in the range of (-1,1).} The normalized scale difference map is further concatenated to $U$ to indicate the object scale information, and injected to the features with a $1 \times 1$ convolution block. Then the squeeze and excitation based channel attention \cite{hu2018squeeze} is used by first squeezing the global average pooled features and then generate a channel weight to capture the channel-wise dependencies. This process can be expressed as:
\begin{equation}
\label{equ_11}
M_{c} = f_{se} (AvgPool(f_{1\times1} (\mathbb{C}[U,S_{U}])))
\end{equation}
where $\mathbb{C}[ , ]$ denotes the concatenation operation, $f_{1\times1}$ refers to the $1\times1$ standard convolution block to inject the scale information to the feature map and $f_{se}$ denotes the squeeze and excitation block. The channel weight $M_{c}$ is applied to the concatenated feature $U$ to obtain the re-weighted feature $U_{v} = M_{c} \odot U $. By this way, it can selectively emphasize crucial information on the different feature channels of the DcSConv features and conventional convolution features since the weights $M_{c}$ effectively measure the importance of corresponding feature channels according to the scale information.

\begin{figure*}[!t]
\centering
\includegraphics[width=0.78\textwidth]{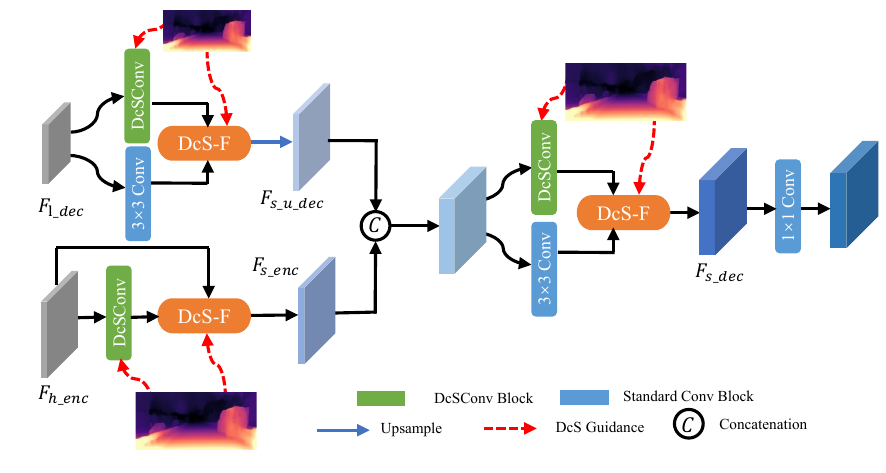}
\caption{Architecture of the Depth-converted-Scale aware feature Decoding (DcS-D) module. Different from the conventional skip-connection based feature decoding, the features from the encoder and the previous decoder layers are first enhanced with the DcSConv and then fused with the conventional convolution features using DcS-F module. }
\label{fig_6}
\end{figure*}

Next, the DcS-spatial attention block fuses the features in a spatial-wise manner to obtain spatial attention weights in order to spatially enhance the DcSConv features against the conventional convolution features based on the scale information at each location. Since the scale information is different at different locations leading to different convolution sizes, the use of scale information in the DcS-spatial attention block is more significant than the DcS-channel attention block. Similar as the DcS-channel attention block, the channel attention enhanced feature $U_{v}$ is first injected with the scale information by concatenating the normalized scale map $S_{U}$ and processing with the $1\times1$ standard convolution block. Then as in the spatial attention of the CBAM \cite{woo2018cbam}, two context descriptors are generated by aggregating the feature with both average pooling and max-pooling in the channel dimension. Then the two descriptors are concatenated and processed by the $7\times7$ standard convolution block to obtain the spatial weight $M_{s}$.  This process can be expressed as:
\begin{align}
\label{equ_12}
&F^\prime = f_{1\times1} (\mathbb{C}[U_{v}, S_{U}])  \nonumber\\
&M_{s} = f_{7\times7} (\mathbb{C}[AvgPool(F^\prime),  MaxPool(F^\prime)]) 
\end{align}
where  $f_{7\times7}$ refers to the $7\times7$ standard convolution block. The spatial weight $M_{s}$ is applied only to the DcSConv features in order to differentiate the spatial importance from the conventional convolution features.

With the DcS-channel and spatial attention, the final feature $F_{s}$ is obtained by fusing the weighted features with a $1\times1$ standard convolution block as:
\begin{equation}
\label{equ_13}
F_{s} = f_{1\times1} (M_{c} \odot \mathbb{C}[M_{s} \odot F_{DcSC}, F_{c}] + U)
\end{equation}
where $\odot$ denotes element-wise dot product. A residual connection is also used to enhance the features, by summing the initial concatenated feature $U$. {Note that here the attention weights are already calculated as in Eqs. (11) and (12), and thus Eq. (13) can be obtained by transforming the order of multiplying the channel and spatial attention weights. This transformation simplifies the expression since the channel attention weight is applied to both the DcSConv feature and general convolution feature, while the spatial attention weight is only applied to the DcSConv feature.}

{The proposed DcS-F can effectively combine the advantages of the DcSConv and general convolution ($3\times 3$) to better capture the features of different objects in the same frame and among different frames. For example, when the object is far away, a smaller-scale convolution is used with DcSConv and thus object details can be extracted, producing sharp edges. When the object is close, a larger-scale convolution is used with DcSConv and thus global information of the object can be better extracted. In such a case, the general $3\times 3$ convolution can capture the details with the guide of object global information using the proposed DcS-F module.}

\subsection{Depth-converted-Scale aware Monocular Depth Estimation}
\label{sec_dcsmde}
As described in {the subsection} \ref{sec_ow} Overview, generally an encoder-decoder architecture is used for the monocular depth estimation. This subsection presents the proposed decoder and the training of the whole framework.

\subsubsection{Depth–converted-Scale aware feature Decoding (DcS-D) module} 

To produce the final depth map, a Depth-converted-Scale aware feature Decoding (DcS-D) module is developed based on the proposed DcSConv and DcS-F, to decode the features from both the encoder and decoder with depth-converted-scale information. It is used as the basic feature decoder at each level to complete the encoder-decoder framework as shown in Fig. \ref{fig_1}. The structure of the proposed DcS-D is illustrated in Fig. \ref{fig_6}. It consists of the further processing of the decoder features from the previous level and the encoder features at the corresponding level, and the processing of the fused features. Each process uses the DcS-F block to fuse the conventional convolution features and the DcSConv features.

Specifically, the proposed DcSConv and the standard ${3\times3}$ convolution are used to process the low-resolution decoded feature from the previous level, in order to first obtain the depth-converted-scale features and general features. Then the proposed DcS-F block is used to obtain the first-stage scale-aware decoded feature $F_{s\_fdec}$. This process can be formulated as
\begin{equation}
\label{equ_14}
F_{s\_fdec} = f_{DcS-F} (\mathbb{C}[f_{DcSC}, f_{Conv}(F_{l\_dec})], S_{l\_dec})
\end{equation}
where $\mathbb{C}[f_{DcSC}, f_{Conv}()]$ represents the concatenated processing of the proposed DcSConv and the standard ${3\times3}$ convolution block, and $F_{l\_dec}$ represents the low-resolution decoder features. $S_{l\_dec}$ is the low-resolution depth-converted-scale map used to guide the DcS-F module ($f_{DcS-F}$) to fuse the DcSConv and general convolution features.

A similar processing is also applied to the high-resolution encoder feature $F_{h\_enc}$ to obtain the scale-aware encoder feature $F_{s\_enc}$:
\begin{equation}
\label{equ_15}
F_{s\_enc} = f_{DcS-F} (\mathbb{C}[f_{DcSC}(F_{h\_enc}),F_{h\_enc}], S_{h\_dec})
\end{equation}
where $S_{h\_dec}$ is the high-resolution depth-converted-scale map and can be obtained by upsampling the low-resolution depth map or an initial high-resolution depth map as discussed in {the subsection} \ref{sec_ow} Overview. Here the encoder feature $F_{h\_enc}$ is already processed with the encoder and thus only enhanced with the DcSConv. The first stage low-resolution scale-aware decoded feature $F_{s\_fdec}$ is then upsampled to $F_{s\_u\_dec}$ and fused with the high-resolution scale-aware encoder feature $F_{s\_enc}$ to obtain the high-resolution decoded feature $F_{s\_dec}$. Similar to generation of  $F_{s\_fdec}$ and $F_{s\_enc}$, $F_{s\_dec}$ is also decoded in a scale-aware way as:
\begin{align}
& F^{\prime\prime} = \mathbb{C}[F_{s\_u\_dec}, F_{s\_enc}]  \nonumber\\
&F_{s\_dec} =  f_{DcS-F} (\mathbb{C}[f_{DcSC}, f_{Conv}(F^{\prime\prime})], S_{h\_dec})
\end{align}
Finally, the decoded feature is processed with a ${1\times1}$ standard convolution to squeeze the feature channels, and then further used to generate the depth map at the corresponding resolution including the final output depth map. 

{
\begin{algorithm}
\label{algorithm_1}
{
	\SetAlgoLined
	\SetKwInOut{Input}{Input}
	\SetKwInOut{Output}{Output}
	\Input{encoder features $ {F_{h\_enc}}$, prior depth map $ {D_{prior}}$.} \Output{depth map $ {D}$.}

	 $ {F_{dec}^4}=  {Conv}( {F_{h\_enc}^4})$; \\
	 \tcp{Process the last encoder feature to the decoder.}

	$depth_{pred}^4={feature\_to\_depth}( {F_{dec}^4})$;\\ 
	\tcp{Predict the depth map at the first scale.} 
	
	\For{$l \gets 3$ \KwTo $0$ \textbf{step} $-1$ }{  
		\tcp{Process decoder features and progressively predict depth maps at multiple scales.}
    	
		 $S_{l\_dec}^l=depth_{pred}^{l+1}$ \textbf{if} $ D_{prior}$  is None \textbf{else} $D_{prior}^{l+1}$;     \\
		 \tcp{Obtain the scale information for the following DcS processing. If prior depth map exists, use prior depth map. Otherwise use the decoded depth map from the previous decoder layer.}

		$F_{s\_fdec}^{l}= {DcS-F}\left(\mathbb{C}\left[ {DcSC}\left( {F_{dec }^{l+1}}\right),  {Conv}\left( {F_{dec}^{l+1}}\right)\right], S_{{dec}}^{l}\right)$;   \\
		\tcp{Obtain the scale aware decoder feature by fusing the features obtained with the DcSConv and general convolution.}

		$S_{h\_dec}^{l}={upsample}(depth_{pred}^{l+1})$ \textbf{if} $ D_{prior}$ is None \textbf{else} $D_{prior}^l$;     \\
		\tcp{Upsample the scale information.}

		$F_{s\_enc}^l= {DCS-F}(\mathbb{C}[ {DcSC} {(F_{h\_enc}^l}), {F_{h\_enc}^l}],S_{h\_dec}^{l}
		)$;   \\
		\tcp{Enhance the encoder feature by fusing the scale aware feature processed with the DcSConv.}
		
		$F^{''}=\mathbb{C}[ {upsample}(F_{s\_fdec}^l),F_{s\_enc}^l]$;  \\
		\tcp{Concatenate the upsampled decoder feature and encoder feature.}
	
		$F_{s\_dec}^l= {DcS-F}(\mathbb{C}[ {DcSC}(F^{''}), {Conv}(F^{''})],S_{h\_dec}^l)$;  \\
		\tcp{Process the encoder and decoder features by fusing the DcSConv and general convolution with DcS-F.}
		
		$depth_{pred}^l= {feature\_to\_depth}(F_{s\_dec}^l)$; 	
		\tcp{Predict the depth map at the current scale.}
		
	} 	
	
    	$ {D}=depth^0_{pred}$; \\
    	\tcp{The final depth map is the predicted depth at the last scale.}
	
	\caption{Depth–converted-Scale aware feature Decoding (DcS-D) module}}
\end{algorithm}}

{The pseudocode of the proposed DcS-D module can be summarized as in algorithm \ref{algorithm_1}.} Our DcS-D module does not concern the details of the network. Therefore, we can easily apply our module as a plug-and-play module to other networks with encoder-decoder framework to further improve the performance of depth estimation.

\begin{table*}[!t]
\setlength{\tabcolsep}{8pt}
\caption{Comparison of different convolution methods based on the Monodepth2 (MD 2) [7]. The best results are shown in bold.
\label{tab:table1}}
\centering
\begin{tabular}{l c c c c c c c}
\hline

Evaluation metric &{Para} & Abs Rel ↓ & Sq Rel ↓ &{$\Delta$Sq Rel↑}  & RMSE ↓  & $\delta < 1.25$ ↑   & $\delta < {1.25^2}$ ↑  \\
\hline
Baseline(MD2)       &{14.34M}          & 0.115	& 0.903  &{/}  & 4.863    & 0.877    & 0.959   \\
Baseline + DMSF      &{25.99M}       & 0.112	 & 0.831 &{7.9\%}  & 4.759    & 0.880    & 0.960    \\
Baseline + Fixed-scale conv  &{17.52M}  & 0.116 & 0.864  	&{4.3\%} & 4.814   & 0.875    & 0.959    \\
Baseline + Deformable conv  &{17.69M} & 0.114 & 0.918 	&{-1.6\%}  & 4.879    & 0.879    & 0.958    \\
Baseline + GLSConv (proposed)         &{18.30M} & 0.112	 & 0.823 &{8.8\%}  & 4.789    & 0.879    & 0.960    \\
Baseline + DcSConv (proposed) &{17.52M} & \textbf{0.112} & \textbf{0.811} &{\textbf{10.2\%}}  & \textbf{4.732}  & \textbf{0.879} & \textbf{0.960}   
 \\
\hline
\end{tabular}
\vspace{-0.2cm}
\end{table*}

\begin{table*}[!t]
	\setlength{\tabcolsep}{8pt}
	\caption{Comparison of different feature fusion methods. The best results are shown in bold.
		\label{tab:table2}}
	\centering
	\begin{tabular}{l c c c c c c c}
		\hline
		
		Evaluation metric &{Para}   &Abs Rel ↓ & Sq Rel ↓ &{$\Delta$Sq Rel  ↑} & RMSE ↓  & $\delta < 1.25$ ↑   & $\delta < {1.25^2}$ ↑  \\
		\hline
		Baseline              &{14.34M}    & 0.115 & 0.903  &{/}  	   & 4.863    & 0.877    & 0.959    \\
		Concatenation          &{17.52M}   & 0.112	& 0.811  &{10.2\%}     & 4.732       & 0.879    & 0.960    \\
		Summation              &{15.75M}   & 0.113	 & 0.824 &{8.7\%}      & 4.754        & 0.878    & 0.960    \\
		CBAM \cite{woo2018cbam}   &{17.57M}  & 0.114	& 0.824 &{8.7\%}      & 4.764        & 0.877    & 0.959    \\
		DcS-F  &{18.42M} &\textbf{0.111} & \textbf{0.798} &{\textbf{11.6\%}}    & \textbf{4.701}  & \textbf{0.882} & \textbf{0.961} \\
		\hline
	\end{tabular}
\vspace{-0.3cm}
\end{table*}

\begin{table*}[!ht]
	\caption{{Results of proposed method in comparison with the baseline on Make3D  and NYUv2. }} 
		\label{tab:nyu_make3d}
	\centering
	{
	\begin{tabular}{l c c c c c c }
		\hline
		
		Method  & Dataset & Abs Rel ↓ & Sq Rel ↓ &$\Delta$Sq Rel ↑   & RMSE ↓  \\ \hline
		Monodepth2~\cite{godard2019digging} & \multirow{2}*{Make3D} &  0.322 & 3.589 & - & 7.417 \\
		Proposed+Monodepth2 & & 0.309 & 3.137 & 12.59\% & 7.083 \\ \hline
		Monodepth2~\cite{godard2019digging} & \multirow{2}*{NYU V2} &  0.384 & 0.753 & - & 1.276 \\
		Proposed+Monodepth2 &  & 0.328 & 0.617 & 18.06\% & 1.228 \\ \hline

	\end{tabular}
	}
\end{table*}

\begin{figure*}[!t]
	\centering
	\includegraphics[width=0.8\textwidth]{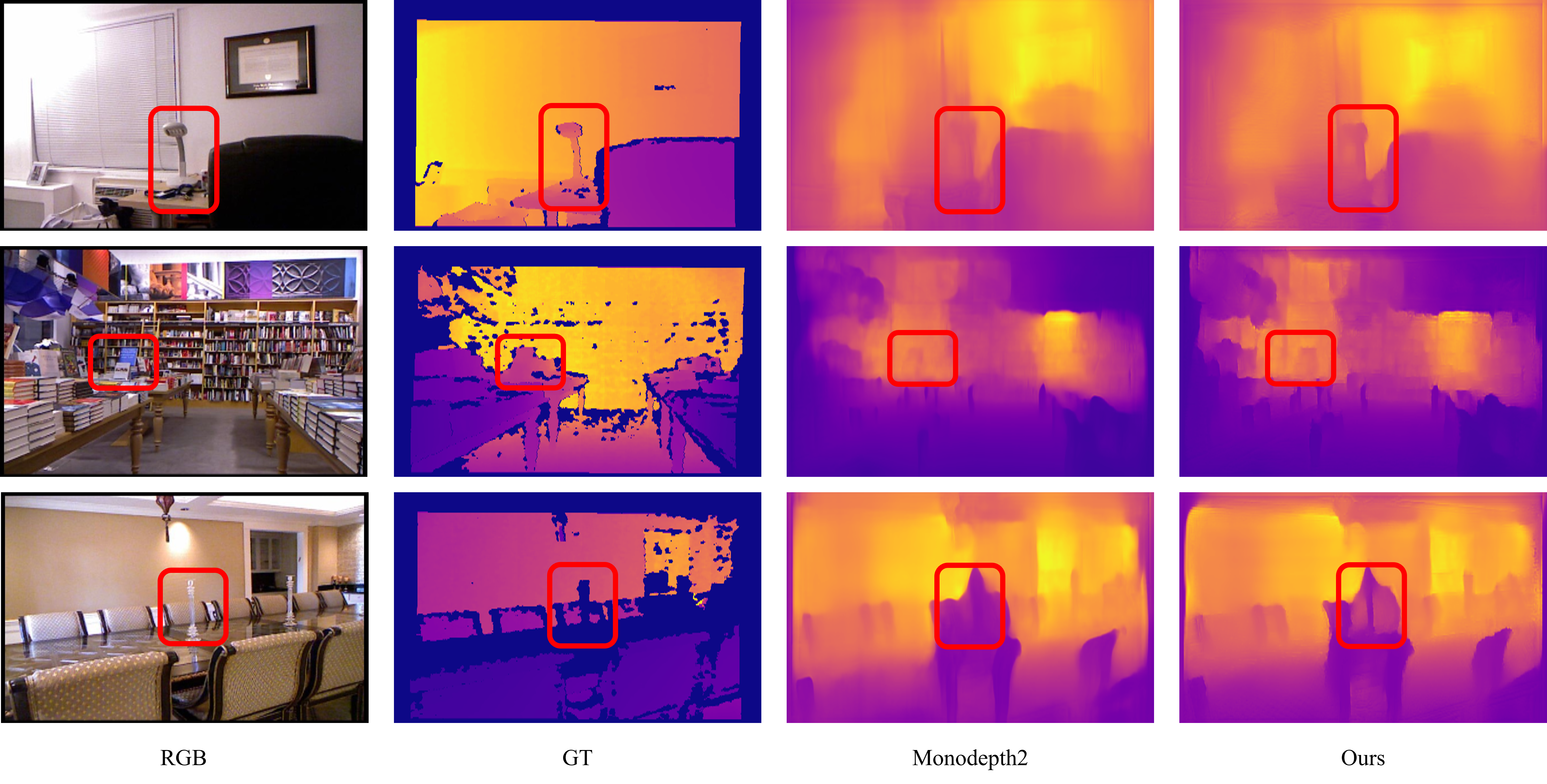}
	\caption{Sample qualitative results from the NYU V2 indoor benchmark dataset.}
	\label{NYUV2}
	\vspace{-0.3cm}
\end{figure*}

\subsubsection{Implementation of the proposed DcSConv in monocular depth estimation as a plug-and-play module} 

As mentioned above, the proposed method can be applied as a plug-and-play module on top of existing self-supervised monocular depth estimation methods. Here, two baseline methods are used including the Monodepth2 \cite{godard2019digging},CADepth \cite{yan2021channel} and MonoViT \cite{zhao2022monovit} to construct the overall network. To be fair for comparison, only the decoding layers are replaced with our DcS-D module. The whole method is trained in an end-to-end manner. Due to page limitations, the network training details are shown in the supplementary material.

\section{Experiments}
\label{sec_exp}
In this {section}, extensive experiments are conducted to evaluate the performance of our method. In the following, we firstly briefly describe the benchmark dataset and the related evaluation metrics, and then introduce our experimental details. Finally, ablation studies on different modules are performed with analysis, and comparison to the state-of-the-art self-supervised monocular depth estimation methods is also provided to demonstrate its effectiveness.
\subsection{Datasets and evaluation metrics}
The KITTI dataset \cite{geiger2012we} has been widely used as benchmark for depth estimation, which is captured in urban, rural and highway scenes. The Eigen split \cite{eigen2015predicting} and the pro-processing in Zhou et al. \cite{zhou2017unsupervised} to remove static frames before training is adopted.   The 56 scenes from the ``city'', ``residential'', and ``road'' categories of the raw data are used, and further split into 28 for training and 28 for testing. Eventually, $39,910$ and $4,424$ monocular image triplets were used for training and validation, respectively, and $697$ images were used for evaluation. We train our network with the random crop of resolution $640\times192$ and evaluate the performance of depth prediction with the per-image median ground truth scaling as in \cite{godard2019digging, yan2021channel}.

The five standard metrics in previous works \cite{godard2019digging,yan2021channel} are used to evaluate our model and compare with state-of-the-art methods, including the Absolute Relative Error (Abs Rel), Square Relative Error (Sq Rel), Root Mean Square Error (RMSE), RMSE log, and threshold accuracy ($\delta$).

\subsection{Implementation Details} 
As described in {the subsection} III.E, our method can be implemented on various existing architectures and the Monodepth2 \cite{godard2019digging}, CADepth \cite{yan2021channel} networks and MonoViT \cite{zhao2022monovit}  are used as the backbone in the experiments. When applying the proposed method on top of existing baseline methods, only the decoder is replaced with our DcS-D module while the encoder is not {changed}. The decoders generally contain five decoding modules.The first decoding module is not changed and directly estimates the first-level depth map. The other four decoding modules are replaced with our DcS-D module using the first-level depth map or the progressively estimated depth map from its previous level. For our GLSConv, no depth map is required and the scale information is directly learned as latent variable. To accelerate training, a multi-stage training strategy is used where the depth maps estimated from the pretrained model are used for the first $10$ training epochs and then the depth maps generated in the decoding architecture as mentioned above are used for subsequent training epochs. {All experiments were implemented in PyTorch and accelerated by an NVIDIA GeForce RTX 3090 GPU. }

\begin{table*}[!ht]
\setlength{\tabcolsep}{10pt}
\caption{Experimental results on the self-supervised monocular depth estimation in comparison with the state-of-the-art methods using KITTI Eigen split. All methods are trained with monocular images of resolution $640\times192$. Improvement percentage in terms of Sq Rel over the Monodepth2 is used as an extra measurement.  The best method is shown in bold for each category of backbone.  
\label{tab:resultall}}
\centering
\begin{tabular}{l l c c c c c c c}
\hline

Method & Backbone  & Abs Rel ↓  & Sq Rel ↓  &{$\Delta$Sq Rel↑} & RMSE ↓  & $\delta < 1.25$ ↑   & $\delta < {1.25^2}$ ↑   \\
\hline
Monodepth2 \cite{godard2019digging} (ICCV) &ResNet 18       & 0.115	  & 0.903 &{/}  & 4.863       & 0.877    & 0.959      \\

{Proposed+Monodepth2} &{ResNet 18}  &{0.112}  &{0.798} &{11.6\%} & {4.701}  & {0.882} & {0.961}   \\

Zhou et al \cite{zhou2019unsupervised} (ICCV) &ResNet 18     & 0.121	   & 0.837 &{7.3\%}  & 4.945        & 0.853    & 0.955   \\

Sun et al \cite{sun2021unsupervised} (TNNLS)&ResNet 18       & 0.117   & 0.863  &{4.4\%} & 4.813    & 0.871    & 0.959      \\

Guizilini et al \cite{guizilini2020semantically} (ICLR) &ResNet 18       & 0.117  & 0.854  &{5.4\%}  & 4.714        & 0.873    & 0.963      \\

SGDepth \cite{klingner2020self} (ECCV) &ResNet 18    & 0.113	  & 0.835 &{7.5\%}  & 4.693    & 0.879    & 0.961   \\

Lee et al. \cite{lee2021learning} (AAAI) &ResNet 18       & 0.112    & 0.777 &{13.9\%}  & 4.772        & 0.880    & 0.962      \\

Zhang et al. \cite{zhang2022self} (TIP) &ResNet 18       & 0.112   & 0.856  &{5.2\%} & 4.778        & 0.880    & 0.961      \\

VC-Depth \cite{zhou2020constant} (CVMP) &ResNet 18        & 0.112   & 0.816 &{9.6\%}  & 4.715        & 0.880    & 0.960   \\

Poggi et al \cite{poggi2020uncertainty} (CVPR) &ResNet 18        & 0.111	  & 0.863 &{4.4\%}  & 4.756        & 0.881    & 0.961   \\

Patil et al \cite{patil2020don} (RAL) &ResNet 18      & 0.111	  & 0.821 &{9.1\%}  & 4.650        & 0.883    & 0.961   \\

HRDepth \cite{lyu2021hr} (AAAI) &ResNet 18          & 0.109   & 0.792 &{12.2\% } & 4.632        & 0.884    & 0.962      \\

Wang et al. \cite{wang2021can} (ICCV) &ResNet 18       & 0.109   & 0.779 &{13.7\%}  & 4.641        & 0.883    & 0.962      \\

CADepth \cite{yan2021channel} (3DV) &ResNet 18          &{0.110}    &{0.812} &{10.1\% } &{4.686}        &{0.882}    &{0.962}      \\

\textbf{Proposed+CADepth} & ResNet 18  & \textbf{0.107}  & \textbf{0.748} &{\textbf{17.1}\%} & \textbf{4.614}  & \textbf{0.884} & \textbf{0.962}   \\
\hline

Guizilini et al \cite{guizilini2020semantically} (ICLR) &ResNet 50       & 0.113   & 0.831 &{8.0\%}  & 4.663        & 0.878    & \textbf{0.971}   \\

SGDepth \cite{klingner2020self} (ECCV) &ResNet 50       & 0.112	  & 0.833 &{7.7\%}  & 4.688        & 0.884    & 0.961   \\

Monodepth2 \cite{godard2019digging} (ICCV) &ResNet 50      & 0.110 	  & 0.831  &{8.0\%}  & 4.642       & 0.883    & 0.962      \\

Wang et al. \cite{wang2020self} (ICASSP) &ResNet 50       & 0.106	  & 0.799 &{11.5\%}  & 4.662      & 0.889    & 0.961      \\

CADepth \cite{yan2021channel} (3DV) &ResNet 50          & 0.105   & 0.769  &{14.8\%}  & 4.535       & 0.892    & 0.964      \\

\textbf{Proposed+CADepth} & ResNet 50   & \textbf{0.103}  & \textbf{0.741} &{\textbf{17.9\%}} & \textbf{4.515}   & \textbf{0.893} & 0.964   \\

\hline

MonoViT \cite{zhao2022monovit} (3DV) &Transformer   & 0.099  & 0.710 &{21.4\%} & 4.391  & 0.900 & 0.967   \\
{Lite-mono \cite{Zhang_2023_CVPR} (CVPR)} & {CNN \& Transformer}   & {0.101}  & {0.729} &{{19.3\%}} & {4.454}  & {0.897} & {0.965}   \\

{\textbf{Proposed+MonoViT}}  &{Transformer}   & {\textbf{0.097}}  & {\textbf{0.692}} &\textbf{23.4\%} & {\textbf{4.373}}   & {\textbf{0.900}} &{\textbf{0.967}}   \\
\hline
\end{tabular}
\vspace{-0.3cm}
\end{table*}
 
\subsection{Ablation Studies}
In this subsection, ablation experiments are conducted based on the Monodepth2 (MD2) \cite{godard2019digging} baseline with ResNet 18 as the encoder to demonstrate the effectiveness of our method. 

\textbf{Evaluation of the effectiveness of Depth-converted Scale Convolution (DcSConv):} In order to validate that our DcSConv can capture better features with different object scales among successive frames, different types of convolutions are evaluated including the baseline MD2, baseline with extra standard $3\times3$ convolution (fixed-scale convolution), the deformable convolution, the proposed DMSF, GLSConv and DcSConv. For the baseline with the fixed-scale convolution, only the DcSConv in our method is replaced with the standard $3\times3$ convolution to maintain the same architecture with similar number of parameters. Similarly for the baseline with the deformable convolution and GLSConv, the convolution in the DcSConv branch is replaced with the deformable convolution and GLSConv, respectively, both learned from a convolution block. { For fair comparison, concatenation is used to fuse the different convolution features for all the above ablation experiments.} The results are listed in Table~\ref{tab:table1}. Several observations and remarks can be drawn from the results. Firstly, it can be seen that the proposed DMSF, GLSConv and DcSConv all outperform the baseline and the baseline with the extra fixed-scale convolution. This verifies our argument that processing the objects with scales from its corresponding depth can help improve the feature extraction. Moreover, the baseline with the fixed-scale convolution of a comparable number of parameters shows no obvious improvement compared to the baseline which can effectively eliminate the impact of network complexity and further prove the effectiveness of the proposed adaptive scale convolution. Secondly, compared with the DMSF, DcSConv achieves better performance, especially in terms of Sq Rel and RMSE. This further demonstrates the advantage of the proposed DcSConv which directly produces a flexible-scale convolution feature based on the depth of the pixels.  Thirdly, by comparing the performance of the GLSConv with the DcSConv, it can be seen that DcSConv outperforms the GLSConv, validating our analysis on the relationship between the depth and scale for the convolution. Lastly, our DcSConv outperforms the baseline with the deformable convolution, proving our argument that the scale of the convolution also matters significantly and even more than the local deformation of the filter shape in this task. 

\begin{figure*}[!t]
\setlength{\abovecaptionskip}{-0.3cm}
\centering
\includegraphics[width=7in]{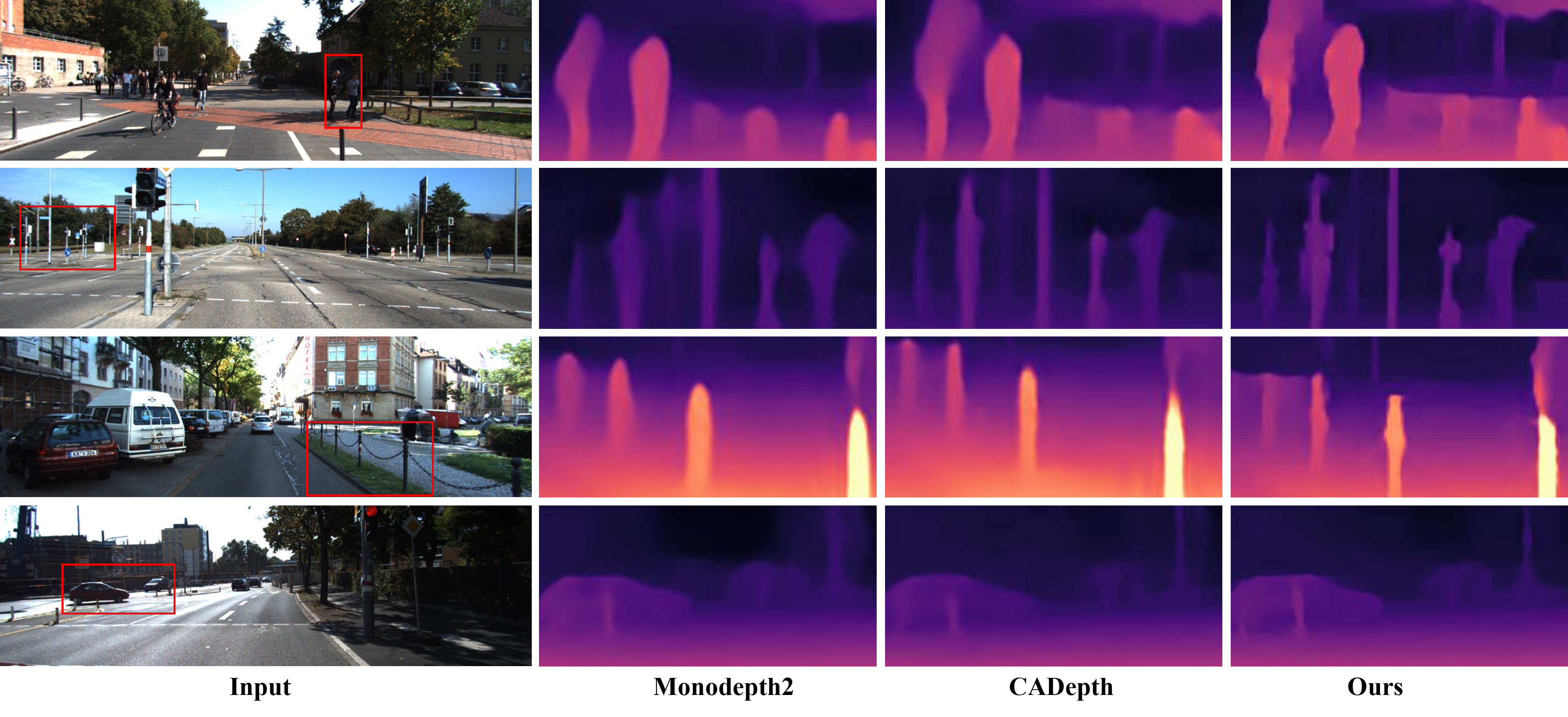}
\caption{{Qualitative compassion of different methods on the KITTI Eigen split. Our model produces better quality depth maps with clearer object edges. }}
\label{fig_8}
\vspace{-0.3cm}
\end{figure*}

\textbf{Evaluation of different feature fusion methods:} In order to validate the effectiveness of the proposed Depth-converted-Scale aware Fusion (DcS-F) module, it is compared with different feature fusion strategies, including concatenation, element-wise summation, and the convolutional block attention module (CBAM) \cite{woo2018cbam}. Compared with other methods, DcS-F combines the flexible-scale convolution features and the fixed-scale convolution features based on the scale information to better extract the scene information. The results are shown in Table~\ref{tab:table2}. First, it can be seen that by combining our DcSConv features using either method, the performances are all greatly improved, further demonstrating the effectiveness of our DcSConv. Moreover, compared to the simple feature combination operations like element-wise summation or concatenation which lack of meticulous attention to local details, our DcS-F module performs the best, up to 11.6\% performance improvement in terms of SqRel reduction with an acceptable increase of parameters. {Especially compared with the CBAM based on the channel and spatial attention, DcS-F greatly improves the performance over the baseline while CBAM only brings a small improvement, validating that the depth/scale information can effectively guide the feature fusion at the channel and spatial dimensions.}

{\textbf{Evaluation of generalization capability on different datasets: }  To further evaluate the generalization capability of the proposed algorithm, experiments on the widely used benchmark datasets other than KITTI, including Make3D~\cite{saxena2008make3d} and NYU V2 \cite{silberman2012indoor}, have also been conducted. Make3D serves as a typical outdoor dataset, while NYU V2 is an indoor dataset, together providing a comprehensive evaluation of the model generalization capability. In this experiment, a pretrained model derived from the KITTI dataset is used. The results are shown in Table \ref{tab:nyu_make3d}. It can be seen that our method achieves performance improvements of 12.59\% and 18.06\%, respectively, over the baseline model (Monodepth2) on these datasets, demonstrating the effectiveness of our approach. Moreover, some sample qualitative results from the NYU V2 indoor benchmark dataset are also illustrated as shown in Fig. ~\ref{NYUV2}. It can be seen that the proposed method achieves more accurate depth estimation at the different objects compared to the baseline.

}

Due to page limitations, other ablation studies including evaluation of different reference depth values, evaluation of using depth information of different qualities for the depth-scale conversion relationship, evaluation of the proposed method as a plug-and-play module are shown in the supplementary material to further demonstrate the effectiveness of our method.

\subsection{Comparison with State-of-the-art Methods}
In this {subsection}, the proposed method is compared with the existing self-supervised monocular depth estimation methods.  The results are shown in Table~\ref{tab:resultall}.  From the results, it can be seen that our method achieves the best results on different backbones including ResNet 18, ResNet 50 and Transformer, and significantly outperforms other networks. It is worth noting that some of the methods also explore additional depth-related auxiliary information, such as semantic labels in \cite{lee2021learning} and scene high-resolution image in \cite{lyu2021hr} while our method only uses the color images. From the results, it can be seen that the performance improvement of the existing methods are all very small, and it is getting even less when the performance reaches a relatively good result such as 0.113 in terms of AbsRel. Compared with the existing methods, the performance improvement of the proposed method is actually significant with notable increase in terms of both AbsRel and SqRel. Specifically, our method on the Monodepth2 improves the performance by 11.6\% in terms of SqRel metrics compared to Monodepth2 baseline, and our method on the CADepth and MonoViT baselines improve the performance up to 17.9\% and 23.4\%, respectively. Even compared with the corresponding baselines, the performance is also improved by 3.6\% and 2.2\%. This better performance over them validates the effectiveness of the proposed DcSConv. {It is worth noting that our method is a general processing method without specific consideration of ground prior such as in \cite{moon2024ground}, and can be used as a plug-and-play module to work on different backbone models to improve the final performance, as shown in the extra ablation studies in the supplementary material.} Fig. \ref{fig_8} shows some qualitative comparison results of the proposed method against the two baseline methods. It can be seen that our method achieves sharper boundaries and fine-grained details as shown in the enlarged region. Taking the estimated depth of the person shown in the first image as an example, the depth estimated by our method clearly reflects the shape of the person with better boundaries than other methods.

\section{Conclusion}
\label{sec_conclusion}
This paper proposes a novel convolution named Depth-converted-Scale Convolution (DcSConv) for monocular depth estimation, to  deal with the continuous object size change among successive video frames  due to the depth change of the object. DcSConv takes advantage of the prior relationship between depth and object size to form adaptive and appropriate scale of the convolution receptive field for processing objects at different depths. Then, a Depth-converted-Scale aware Fusion (DcS-F) module is developed to fuse the DcSConv features and the conventional convolution features by taking the scale information into consideration. The DcSConv and DcS-F can be used in place of the conventional convolution for any existing CNN-based monocular depth estimation architectures to accommodate the scale changes due to the different depths.The DcSConv can also be extended to GLSConv without explicitly using the depth information but with learned latent variable. Ablation studies validate the effectiveness of the proposed DcSConv and indicating that the scale of the convolution filter is more significant than its local deformation in this task. Experimental results on KITTI demonstrate that the proposed DcSConv enhanced monocular depth estimation achieves better performance than the state-of-the-art methods.

\bibliography{mybibfile}
\bibliographystyle{ieeetr} 
\end{document}